\documentclass[runningheads]{llncs}
\usepackage{graphicx}
\usepackage{amsmath,amssymb} 
\usepackage{color}
\usepackage{hyperref}

\begin{document}
\pagestyle{headings}
\mainmatter


\title{EPSNet: Efficient Panoptic Segmentation Network with Cross-layer Attention Fusion} 


\titlerunning{EPSNet: Efficient Panoptic Segmentation Network}
\authorrunning{Chang et al.}

%

%
%

\author{Chia-Yuan Chang\inst{1} \orcidID{0000-0001-5841-3455}\and
Shuo-En Chang\inst{1}\orcidID{0000-0003-4916-499X} \and
Pei-Yung Hsiao \inst{2}\orcidID{0000-0003-1750-7118} \and
Li-Chen Fu \inst{1}\orcidID{0000-0002-6947-7646}
}
\institute{National Taiwan University, Taipei, Taiwan 
\email{\{r07922102,r08922a02,lichen\}@ntu.edu.tw} \and
National University of Kaohsiung, Kaohsiung, Taiwan \\
\email{pyhsiao@nuk.edu.tw}
}

\maketitle

\begin{abstract}
Panoptic segmentation is a scene parsing task which unifies semantic segmentation and instance segmentation into one single task. However, the current state-of-the-art studies did not take too much concern on inference time. In this work, we propose an Efficient Panoptic Segmentation Network  (EPSNet) to tackle the panoptic segmentation tasks with fast inference speed.  Basically, EPSNet generates masks based on simple linear combination of prototype masks and mask coefficients. The light-weight network branches for instance segmentation and semantic segmentation only need to predict mask coefficients and produce masks with the shared prototypes predicted by prototype network branch. Furthermore, to enhance the quality of shared prototypes, we adopt a module called "cross-layer attention fusion module", which aggregates the multi-scale features with attention mechanism helping them capture the long-range dependencies between each other.
To validate the proposed work, we have conducted various experiments on the challenging COCO panoptic dataset, which achieve highly promising performance with significantly faster inference speed (51ms on GPU). Code is available at: \url{github.com/neo85824/epsnet}.


\end{abstract}

\section{Introduction}

Due to Convolutional Neural Networks (CNNs) and other advances in deep learning, computer vision systems have achieved considerable success especially on computer vision tasks such as image recognition \cite{He2016}, semantic segmentation \cite{Zhao2016,Chen2018}, object detection \cite{Ren2017,Redmon2017} and instance segmentation \cite{He2017,Liu}. In particular, semantic segmentation aims to assign specific class label for each image pixel, whereas instance segmentation predicts foreground object masks. However, the former is not capable of separating objects of the same class, and the latter only focuses on segmenting of \emph{things} (i.e countable objects such as people, animals, and tools) rather than \emph{stuff} (i.e amorphous regions such as grass, sky, and road). 
To overcome the respective shortcomings, combination of semantic segmentation and instance segmentation leads to the so-called panoptic segmentation \cite{Li2018}. More specifically, the goal of panoptic segmentation is to assign a semantic label and an instance ID to every pixel in an image.

Several methods \cite{Xiong2019,Sofiiuk2019,Yang2019a,Liu2019,Li2018c,DeGeus2018a} have been proposed for panoptic segmentation in the literature. Detection-based approaches \cite{Xiong2019,Liu2019,Li2018c,DeGeus2018a} usually exploit an instance segmentation network like Mask R-CNN \cite{He2017} as the main stream and attach light-weight semantic segmentation branch after the shared backbone. Then, they combine those outputs by heuristic fusion \cite{Li2018} to generate the final panoptic prediction. Despite such detection-based fashions which achieve the state-of-the-art results, they solely aim to improve the performance but may sacrifice the computation load and speed. In fact, detection-based methods suffer from several limitations. First, due to the two-stage detector, instance segmentation branch costs the major computation time and drags down the inference speed. Second, most detection-based approaches commonly employ the outputs of backbone, like feature pyramid network \cite{Lin2017}, as shared features without further enhancement, causing sub-optimality of features used by the following branches.
Lastly, the independent branches unfortunately lead to inconsistency when generating final prediction.
\begin{figure}[t]
    \begin{minipage}{0.45\textwidth}
    \includegraphics[width=0.9\linewidth, ]{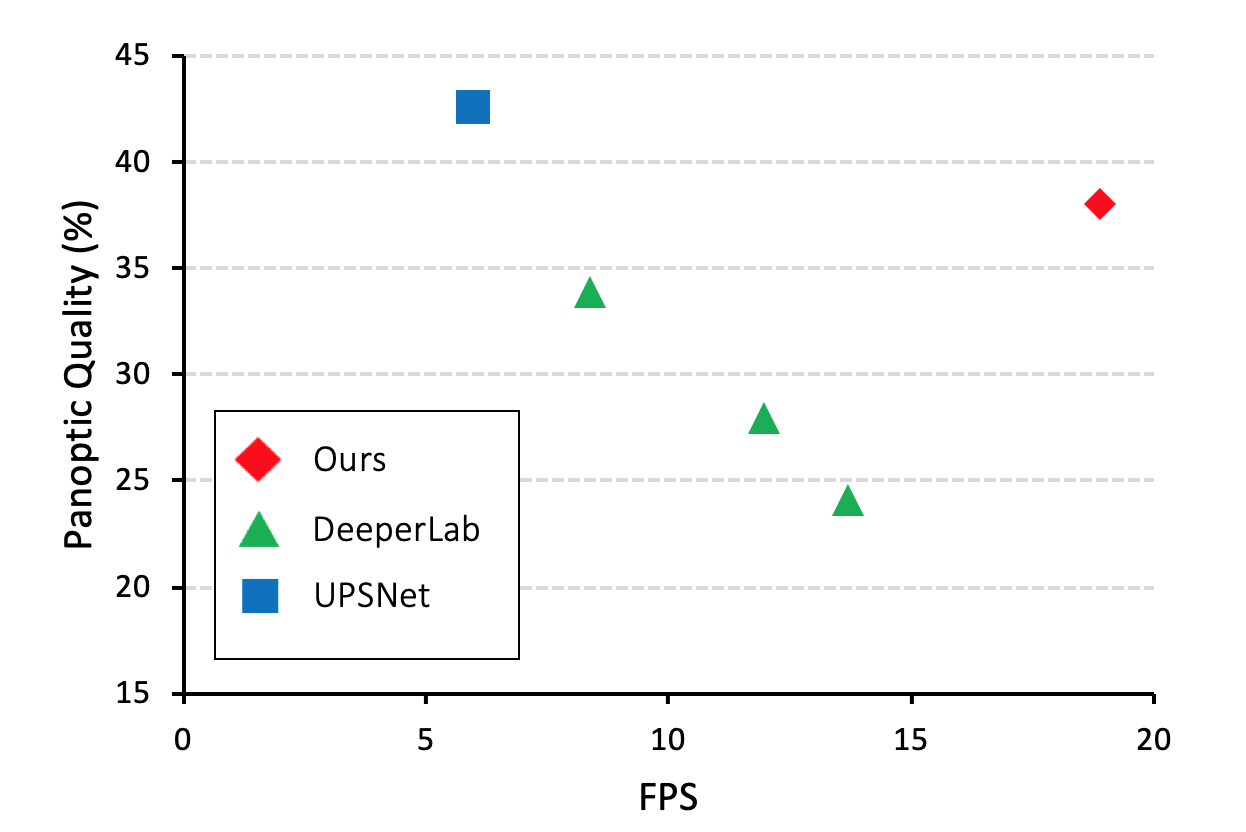}
    \caption{Speed-performance trade-off of panoptic segmentation methods on COCO. The inference time is measured end-to-end from input image to panoptic segmentation output. Our approach achieves 19 fps and 38.6\% PQ on COCO $val$ set. }
    \label{fig:sp_tradeoff}
    \end{minipage}
    \hspace{.02\textwidth}
    \begin{minipage}{0.5\textwidth}
    \includegraphics[width=0.99\linewidth,  ]{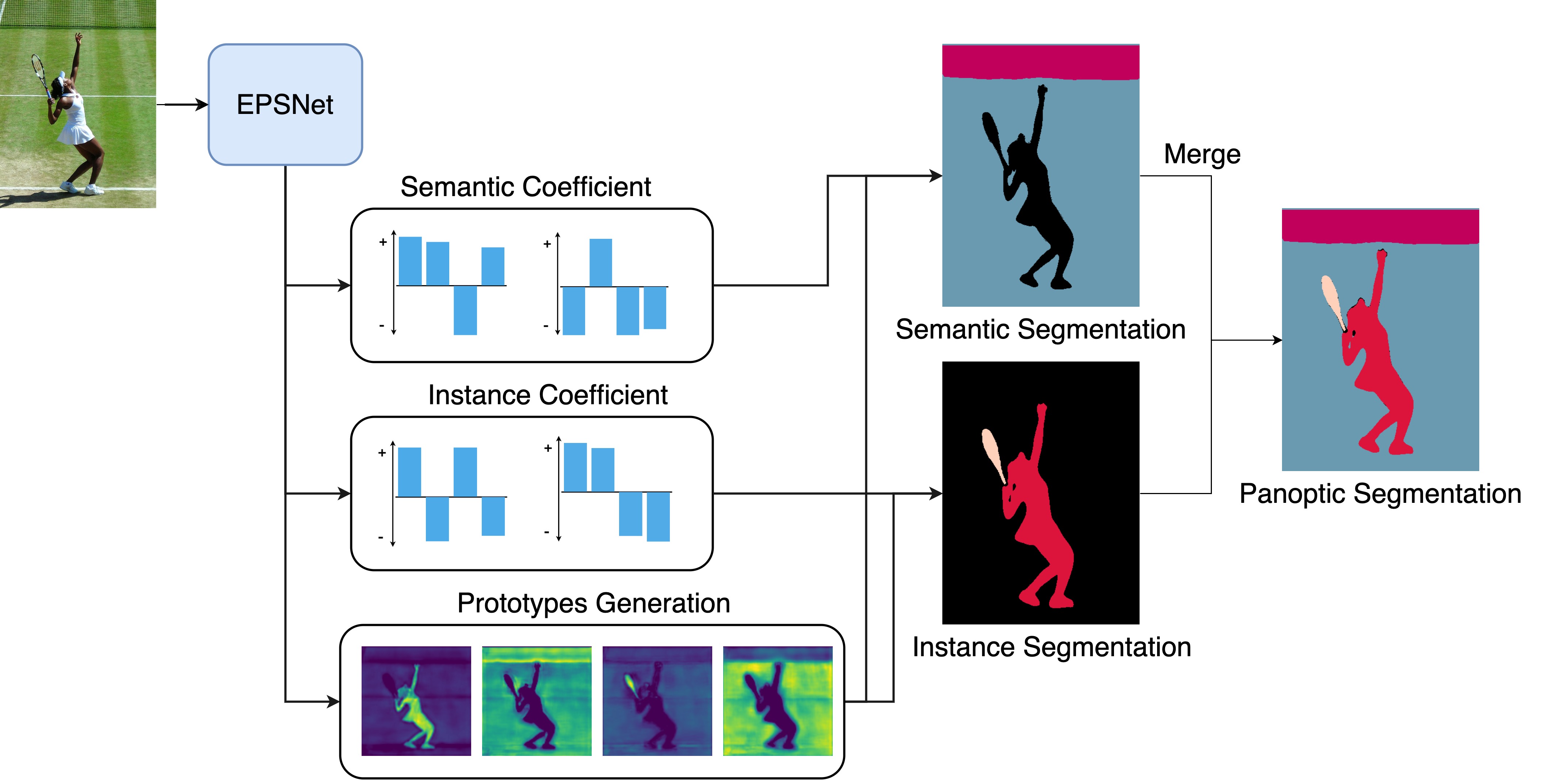}
    \caption{Overview of Efficient Panoptic Network. EPSNet predicts prototypes and mask coefficients for semantic and instance segmentation. Both segmentation, obtained by linear combination of prototypes and mask coefficients, are fused using heuristic merging. }
    \label{fig:simple_overview}
    \end{minipage}
\end{figure}

To address the above problems, we propose a novel one-stage framework called Efficient Panoptic Segmentation Network (EPSNet), as shown in Fig. \ref{fig:simple_overview}. It adopts parallel networks to generate prototype masks for the entire image and predicts a set of coefficients for instance and semantic segmentation. Instance and semantic segments can be easily generated by linearly combining the prototypes with predicted coefficients from the branches. The proposed semantic branch only needs to produce coefficients for each class instead of pixel-wise predictions. Moreover, the prototypes are shared by both branches, which save time for producing large-size masks and help them solve their tasks simultaneously. Further, we introduce an innovative fusion module called cross-layer attention fusion module, which enhances the quality of shared features with attention mechanism. Instead of directly using suboptimal features in FPN, we choose certain layer as the target feature and other layers as source features and then apply an attention module on them to capture spatial dependencies for any two positions of the feature maps. For each position in target feature, it is updated via aggregating source features at all positions with weighted summation.
To verify the efficiency of EPSNet, we conduct experiments on  COCO \cite{Caesar2018} dataset. The experimental results manifest that our method achieves competitive performances with much faster inference compared to current approaches, as shown in Fig. \ref{fig:sp_tradeoff}. 

\section{Related Work}
\subsection{Panoptic Segmentation}
Panoptic segmentation is originally proposed by \cite{Li2018}. In panoptic segmentation tasks, each pixel in the image needs to be assigned a semantic label and an instance ID. In \cite{Li2018}, separate networks are used for semantic segmentation and instance segmentation, respectively, and then the results are combined with heuristic rules.
The recent approaches of panoptic segmentation train semantic and instance segmentation network in end-to-end fashion with shared backbone. These methods can be categorized into two groups, namely, detection-based methods and bottom-up methods. 
\subsubsection{Detection-based.}	
Most detection-based methods exploit Mask R-CNN \cite{He2017} as their instance segmentation network and attach semantic segmentation branch with FCN \cite{Long2015} after shared backbone. These approaches are also considered as two-stage methods because of the additional stage to generate proposals. For instances, JSIS \cite{DeGeus2018} firstly trains instance and semantic segmentation network jointly. TASCNet \cite{Li2018a} ensures the consistency of stuff and thing prediction through binary mask. OANet \cite{Liu2019} uses spatial ranking module to deal with the occlusion problem between the predicted instances. Panoptic FPN \cite{DeGeus2018a} endows Mask R-CNN \cite{He2017} with a semantic segmentation branch. AUNet \cite{Li2018c} adds RPN and thing segmentation mask attentions to stuff branch to provide object-level and pixel- level attentions. UPSNet \cite{Xiong2019} introduces a parameter-free panoptic head which solves the panoptic segmentation via pixel-wise classification. The First  weakly-supervised method is proposed by \cite{Li2018b} to diminish the cost of pixel-level annotations. SOGNet \cite{Yang2019a} encodes overlap relations without direct supervision to solve overlapping. AdaptIS \cite{Sofiiuk2019} adapts to the input point with a help of AdaIN layers \cite{Karras2019} and produce masks for different objects on the same image. Although detection-based methods achieve better performance, they are usually slow in inference because of two-stage Mask R-CNN \cite{He2017} in instance head. In addition, the inconsistency of semantic and instance segmentation needs to be solved when the two are merged into panoptic segmentation.

\subsubsection{Bottom-up.}
Unlike the above approaches, some methods tackle panoptic segmentation tasks by associating pixel-level predictions to each object instance \cite{Zhang2016,Zhang2015,Uhrig2016,Liang2018}. In these approaches, they first predict the foreground mask with semantic segmentation, and then use several types of heatmaps to group foreground pixels into objects.
DeeperLab \cite{Yang2019} predicts instance keypoint as well as multi-range offset heatmap and then groups them into class-agnostic instance segmentation. In semantic segmentation head, they follow the design of DeepLab \cite{Chen2018}. At the end, panoptic segmentation is generated by merging class-agnostic instance masks and semantic output. 
SSAP \cite{Porzi2019} groups pixels based on a pixel-pair affinity pyramid with an efficient graph partition method.
Despite the single-shot architecture of bottom-up approaches, their post-processing step still needs major computational time. Also, the performance of the bottom-up methods usually is inferior to that of the detection-based methods.

Recently, the proposed methods obtain shared feature for semantic and instance head. The quality of shared feature is highly essential for the following network head to produce better results. Still, the proposed approaches do not take this into consideration, and they usually make use of the output of shared backbone as shared feature directly. 

In this work, we aim to propose a panoptic segmentation network based on one-stage detector to attain fast inference speed and competitive performance. To increase the quality of shared feature, our proposed cross-layer attention fusion, which is a lightweight network, provides the target feature map with richer information in different feature pyramid layers using attention mechanism.

\section{Efficient Panoptic Segmentation Network}
\subsection{Efficient Panoptic Segmentation Network}
Our method consists of five major components including (1) shared backbone, (2) protohead for generating prototypes, (3) instance segmentation head, (4) semantic segmentation head, and (5) cross-layer attention fusion module for aggregating multi-scale feature.

\subsubsection{Backbone.}  
\begin{figure}[tbh]
\centering
\includegraphics[width=0.95\linewidth, ]{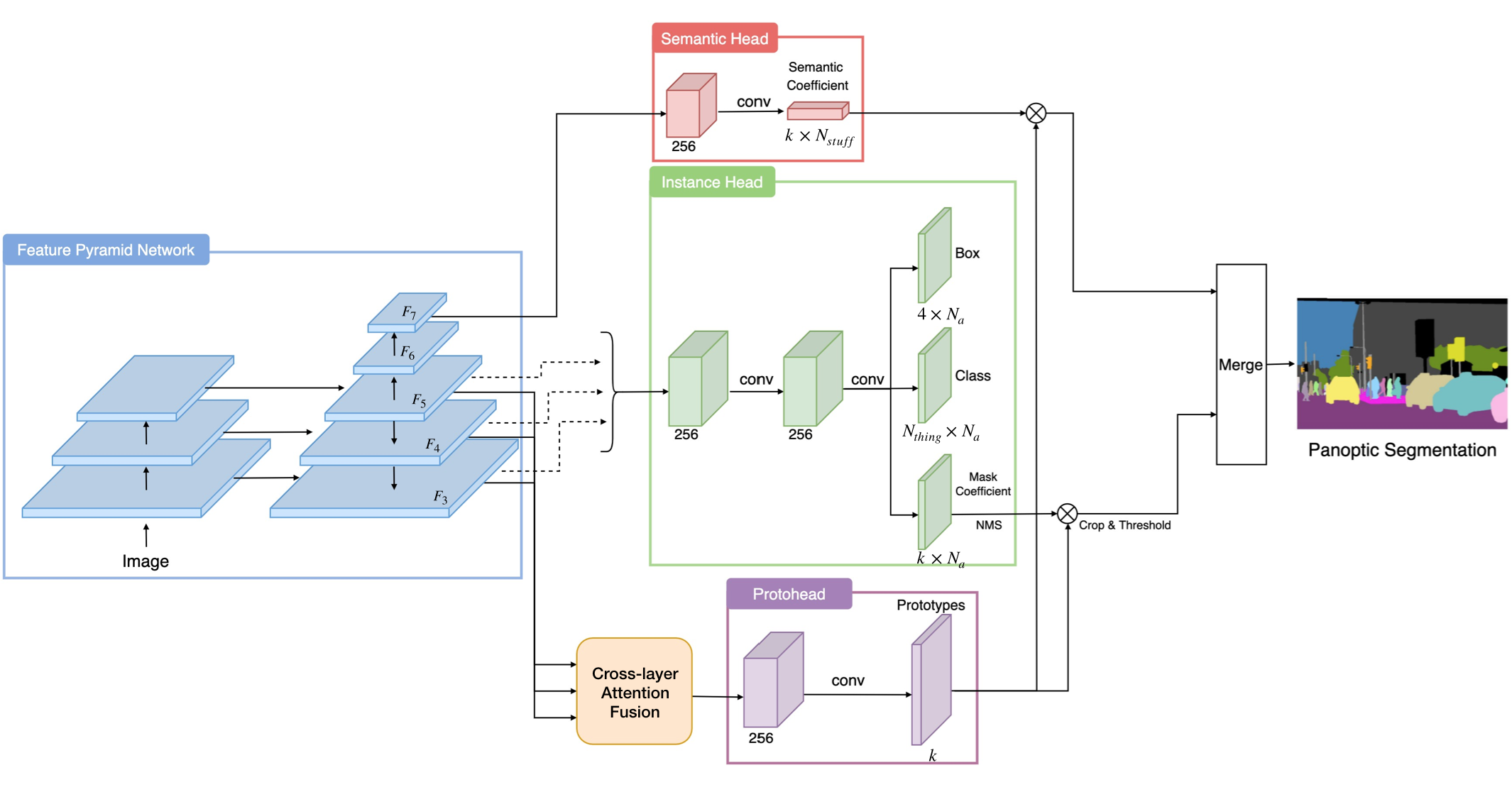}
\caption{Architecture of EPSNet. We adopt ResNet-101 \cite{He2016} with FPN \cite{Lin2017} as our backbone and only exploit $F_3$, $F_4$, $F_5$ for cross-layer attention fusion module. The prototypes are shared by semantic and instance head. $k$ denotes the number of prototypes. $N_a$ denotes the number of anchors. $N_{\text{thing}}$ and $N_{\text{stuff}}$ stands for the number of thing and stuff classes, respectively. $\otimes$ means matrix multiplication. }
\label{fig:overview}
\end{figure}

Our backbone exploits a deep residual network (ResNet) \cite{He2016} with a feature pyramid network (FPN) \cite{Lin2017}, which takes a standard network with features at multiple spatial resolutions and adds a light top-down pathway with lateral connections. It generates a pyramid feature with scales from 1/8 to 1/128 resolution ($F_3$ to $F_7$), where each pyramid channel dimension is set to 256, see Fig. \ref{fig:overview}. For these features, $F_7$ is fed to the semantic head, and $F_3$ to $F_5$ are sent to instance head and protohead as inputs. 


\subsubsection{Protohead.}
Rather than producing masks with FCN \cite{Long2015}, inspired by Yolact \cite{Bolya}, we choose to combine prototypes and mask coefficients with linear combination to generate masks. Our network heads only need to deal with mask coefficients and construct masks with shared prototypes. The goal of protohead is to provide high-quality prototypes which contain semantic information and details of high-resolution feature.


To generate higher resolution prototypes with more semantic values, we perform cross-layer attention fusion module to aggregate multi-scale features in backbone into information-richer feature maps for protohead as inputs. 
Then, we apply three convolutional blocks, $2\times$ bilinear upsampling and $1\times1$ convolution to produce output prototypes which are at 1/4 scale with $k$ channels. 


\subsubsection{Instance Segmentation Head.}

In most panoptic segmentation networks, they adopt Mask R-CNN \cite{He2017} as their instance segmentation branch. Yet, Mask R-CNN \cite{He2017} needs to generate proposals first and then classify and segment those proposals in the second stage. Inspired by one-stage detector, Yolact \cite{Bolya}, our instance head directly predicts object detection results and mask coefficients to make up the segmentation with prototypes without feature localization (e.g. ROI Align \cite{He2017}) and refinement.

The instance head aims to predict box regression, classification confidences and mask coefficients. There are three branches in instance head. Regression branch predicts 4 box regression values, classification branch predicts $N_{\text{thing}}$ class confidences, and mask branch predicts $k$ mask coefficients. Thus, there are totally $4+N_{\text{thing}}+k$ values for each anchor. We perform a convolutional block on input features ($F_3$ to $F_5$) first and send them to each branch to predict respective results. In mask branch, we choose $\tanh$ as the activation function, which allows subtraction when linearly combining the coefficients.



In inference, we choose the mask coefficients whose corresponding bounding boxes survive after NMS procedure. Then, we combine mask coefficients and prototypes  generated from protohead with linear combination followed by $sigmoid$ to produce instance masks and crop final mask with predicted bounding box. During training, we crop mask with ground truth bounding box and divide mask segmentation loss by the area of ground truth bounding box.

\subsubsection{Semantic Segmentation Head.}

\begin{figure}[!tb]
\centering
\includegraphics[width=0.98\linewidth, ]{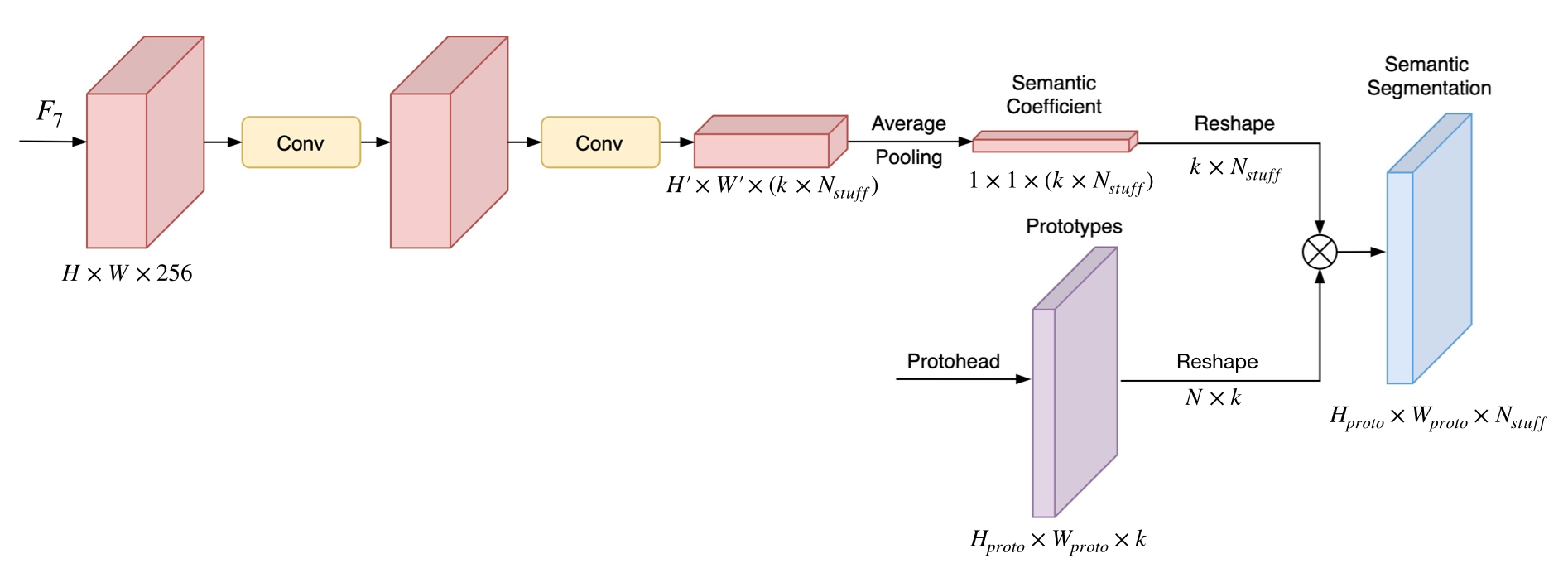}
\caption{The design of proposed semantic head. We adopt $F_7$ as input feature and apply two convolutional blocks on it without padding. The semantic coefficients are produced after average pooling, which predicts $k$ mask coefficients for each stuff classes. Here, $\otimes$ denotes matrix multiplication. }
\label{fig:sem_head}
\end{figure}

Usually, semantic segmentation masks are generated by decoder network \cite{Lin2017a,Zhang2018,Zhao2016,Long2015}, which applies FCN \cite{Long2015} networks and up-sampling layer on the features from backbone to make sure that the size of semantic segmentation outputs is similar to the original input size. However, due to the large feature maps, the computation speed is limited by image size.

To reduce the computation of large feature maps, we propose a novel semantic segmentation head which only produces mask coefficients for each class, see Fig. \ref{fig:sem_head}. The semantic masks can be easily generated by combining the coefficients and prototypes, and each semantic class only demands $k$ mask coefficients. Therefore, with smaller feature maps, the proposed light-weight semantic head can achieve faster inference speed.

We adopt last layer $F_7$ from backbone as the input of semantic head. Two convolution blocks are performed. In the second convolution block, the output channel is set to $k\times N_{\text{stuff}}$ and $\tanh$ is used as activation function. Because of the channel size $k\times N_{\text{stuff}}$, every position in the feature map can predict $k$ coefficients for each class to construct semantic segmentation. Accordingly, we perform average pooling to aggregate the mask coefficients from all positions to generate final semantic coefficients. Further, prototypes from protohead and semantic coefficients are reshaped to 2d matrix and applied with linear combination followed by $softmax$ to produce semantic segmentation result. The operation is able to be implemented by matrix multiplication which is defined as
\begin{equation}
S = softmax(P\cdot Y),
\end{equation}
where $P\in \mathbb{R}^{N\times k}$ denotes prototypes , $Y \in \mathbb{R}^{k\times N_{\text{stuff}}}$ stands for the reshaped semantic coefficients, and $S \in \mathbb{R}^{N \times N_{\text{stuff}}}$ is semantic segmentation result. $N_{\text{stuff}}$ represents the number of stuff classes including 'other' class and $N$ denotes the number of locations in prototypes.

Although the channels of semantic coefficients depend on the number of classes and prototypes, the feature maps in semantic head are much smaller than the large feature maps in other approaches. Our semantic head provides faster semantic segmentation generation and less computation cost.

\subsubsection{Cross-layer Attention Fusion.}
\begin{figure}[!tb]
    \begin{minipage}{.5\textwidth}
    \centering
      \includegraphics[width=.95\linewidth]{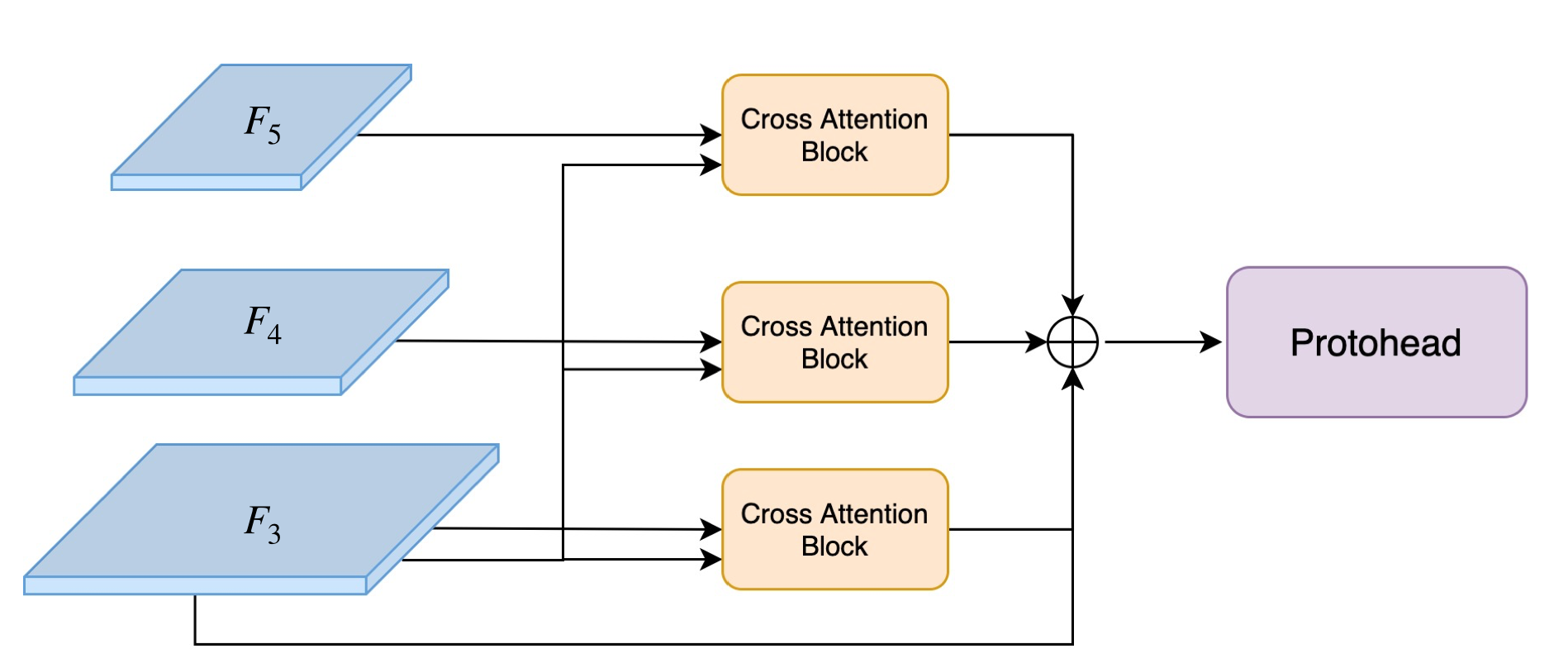}
      \caption{The architecture of cross-layer attention module. The layers $F_3$, $F_4$ and $F_5$ in FPN are used. $F_3$ is considered as target feature, and all of them are set as source features. $\oplus$ denotes element-wise addition. }
      \label{fig:cross_attention_module}
      \centering
    \end{minipage}
    \hspace{.05\textwidth}
    \begin{minipage}{.4\textwidth}
    \centering
      \includegraphics[width=0.9\linewidth]{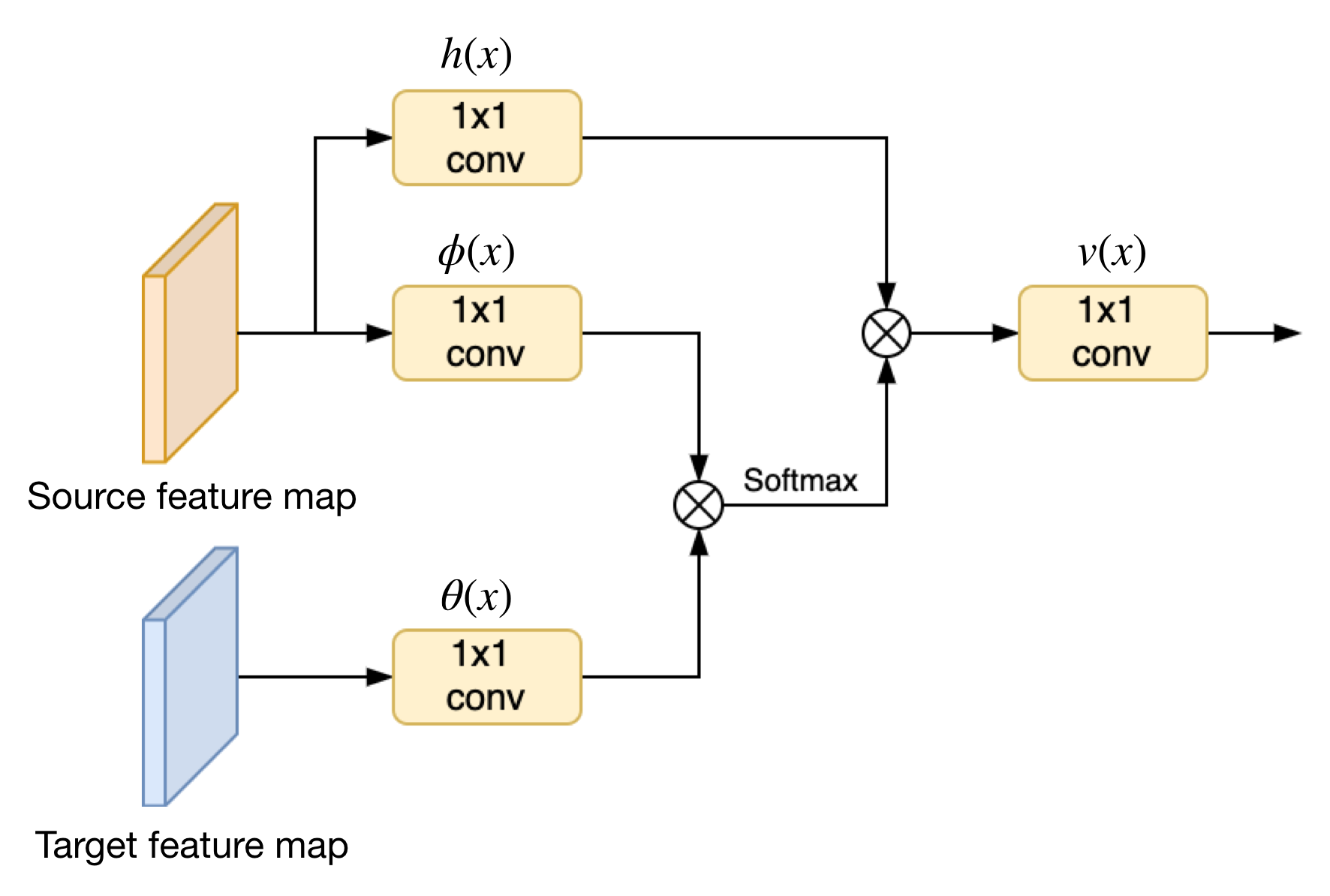}
      \caption{The architecture of cross attention block. $\otimes$ denotes matrix multiplication. }
      \label{fig:cross_attention_block}
    \centering
    \end{minipage}
\end{figure}
Since all of the semantic or instance masks are derived from linear combination of mask coefficients and prototypes which are shared to semantic and instance head, the quality of prototypes significantly influences the generated masks. Hence, we decide to enrich the input features of protohead to help protohead produce preferable prototypes.

To enhance inputs of protohead, we propose a module called Cross-layer Attention (CLA) Fusion Module. This fusion module aims to efficiently aggregate the multi-scale feature maps in FPN layers. Certain layer within the module is chosen to be the target feature map and extract values from source feature maps with attention mechanism, as shown in Fig. \ref{fig:cross_attention_module}. To provide high resolution details, the layer $F_3$, which is the highest resolution feature map in FPN, is selected as target feature map, and we choose other layers (e.g. $F_3$ to $F_5$) as source features to provide more semantic values for target. 

Instead of directly merging multi-scale features with element-wise addition or concatenation \cite{Zhao2016,Xiong2019,Pang,Porzi2019a}, we combine them with a block called Cross Attention Block. Inspired by non-local \cite{Wang2018} and self-attention \cite{Zhang,Fu,Vaswani2017}, which are able to capture long-range dependencies efficiently in feature map, the proposed cross attention block follows their concepts and further finds the long-range relationships from two different feature maps, as shown in Fig. \ref{fig:cross_attention_block}. Each position in target feature map is updated with the weighted sum of all positions in source feature, where attention weight is calculated by similarities between the corresponding positions.

For the target feature map $T \in \mathbb{R}^{C\times N}$ and source feature map $O \in \mathbb{R}^{C\times N}$, we first transform these feature maps into two feature spaces $\theta(x) =W_\theta x$ and $\phi(x) =W_\phi x$ and calculate the attention score $s$ with dot product as shown below
\begin{equation}
   s_{i,j} = \theta(T_i)^T\phi(O_j),
\end{equation}
where $s_{i,j}$ measures the attention score of position $j$ in $O$ and position $i$ in $T$. Here, $C$ denotes the number of channels, and $N$ denotes the number of locations from feature maps. After that, we obtain the attention weight $\alpha$ by normalizing attention score for each position $j$ with $softmax$
\begin{equation}
    \alpha_{i,j} = \frac{\exp(s_{i,j})}{\sum_{j=1}^N{\exp(s_{i,j})}}, 
\end{equation}
where $\alpha_{i,j}$ stands for the normalized impact of position $j$ in source feature map to position $i$ in target feature map. Then, each position in output feature map $A \in \mathbb{R}^{C\times N}$ is produced by calculating weighted sum of source features across all positions. The operation is shown as follows
\begin{equation}
    A_i =v(\sum_{j=1}^N{\alpha_{i,j}h(O_j})).
\end{equation}
Here, $A_i$ denotes output feature on position $i$, and both $v(x) =W_v x $ and $h(x) =W_h x$ stand for embedding functions. The embedding functions $\theta$, $\phi$, $h$ and $v$ are implemented by $1 \times 1 $ convolution, and their output channels are set to 128, which is 1/2 of input channel to reduce computation cost. Finally, we apply cross attention block on each layer including $F_5$, $F_4$ and $F_3$ and consider $F_3$ as target feature map. The overall operation is defined as
\begin{equation}
    Z = A^{F_3,F_5} + A^{F_3,F_4} + A^{F_3,F_3} + F_3,
\end{equation}
where $Z$ denotes the aggregated result from source and target features. Also, we adopt residual connection that makes a new cross attention block easier to insert without interfering the initial behaviors. 
 
With the cross attention block, each position in the target feature is able to obtain spatial dependencies over all positions in feature maps from other layers. Moreover, we also select $F_3$ as source feature map even $F_3$ is target feature. In this case, it is same as self-attention, which helps target feature capture long-dependencies on its own feature map. 

\subsection{Training and Inference}
During training, our EPSNet contains 4 loss functions in total, namely,  classification loss $\mathcal{L}_c$, box regression loss $\mathcal{L}_b$, instance mask segmentation loss $\mathcal{L}_m$ and semantic segmentation loss $\mathcal{L}_s$.
Because each loss function is in different scales and normalized policies, different weights on different loss functions actually affect the final performance on instance and semantic branch. Thus, we set several hyper-parameters on those loss functions, which is defined as $\mathcal{L}=\lambda_c\mathcal{L}_c + \lambda_b\mathcal{L}_b + \lambda_m\mathcal{L}_m + \lambda_s\mathcal{L}_s $.

In inference, since we won’t allow overlaps on each pixel in panoptic segmentation, we resolve overlaps in instance segmentation with post-processing proposed in \cite{Li2018}. 
After getting non-overlapping instance segmentation results, we resolve possible overlaps between instance and semantic segmentation in favor of instance.
Further, the stuff regions which are predicted as 'other' or under a predefined threshold are removed.



\section{Experiments}
In this section, we conduct experiments on COCO \cite{Caesar2018} panoptic dataset. 
Experimental results demonstrate that EPSNet achieves fast and competitive performance on COCO dataset.

\subsection{Experimental Setup}
\subsubsection{Datasets.}
COCO \cite{Caesar2018} panoptic segmentation task consists of 80 thing classes and 53 stuff classes. There are approximately 118K images on training set and 5K on validation set.
\subsubsection{Metrics.}
We adopt the evaluation metric called panoptic quality (PQ), which is introduced by \cite{Li2018}. Panoptic quality is defined as:
\begin{equation}
    PQ = \underbrace{\frac{\sum_{(p,g)\in \text{TP}}\text{IoU}(p,g)}{|\text{TP}|} }_{\text{SQ}}  
    \underbrace{\frac{|\text{TP}|}{|\text{TP}| + \frac{1}{2}|\text{FP}| + \frac{1}{2}|\text{FN}|}}_{\text{RQ}},
\end{equation}
which can be considered as the multiplication of semantic quality (SQ) and recognition quality (RQ). Here, $p$ and $g$ are predicted and ground truth segments. TP, FP and FN represent the set of true positives, false positives and false negatives, respectively. 
\subsubsection{Implementation Details.}
We implement our method based on Pytorch \cite{Paszke2019} with single GPU RTX 2080Ti. The models are trained with batch size 2. Owing to the small batch size, we freeze the batch normalization layers within backbone and add group normalization \cite{Wu2018} layers in each head. The initial learning rate and weight decay are set to $10^{-3}$ and $5\times10^{-4}$. We train with SGD for 3200K iterations and decay the learning rate by a factor of 10 at 1120k, 2400K, 2800k and 3000k iterations and a momentum of 0.9. The loss weights $\lambda_c$, $\lambda_b$, $\lambda_m$ and $\lambda_s$ are 1, 1.5, 6.125 and 2, respectively.

Our models are trained with ResNet-101 \cite{He2016} backbone using FPN with ImageNet \cite{Russakovsky2015} pre-trained weights. We adopt similar training strategies in backbone and instance segmentation head as Yolact \cite{Bolya}. The number of prototypes $k$ is set to 32. Our instance head only predicts thing classes, and semantic head predicts stuff classes viewing thing class as other. The cross attention blocks are shared in cross-layer attention fusion module. The base image size is $550\times550$. We do not preserve aspect ratio in order to get consistent evaluation times per image. We perform random flipping, random cropping and random scaling on images for data augmentation. The image size is randomly scaled in range $[550, 825]$ and then randomly cropped into $550 \times550$.

\subsection{Ablation Study}

\begin{table}[!tb]
\centering
\caption{Ablation study on COCO panoptic \emph{val} set with panoptic quality (PQ), semantic quality (SQ) and recognition quality  (RQ). PQ$^{\text{Th}}$ and PQ$^{\text{St}}$ indicate PQ on thing and stuff classes. \emph{Data Aug} denotes data augmentation. \emph{CLA Fusion} stands for cross-layer attention fusion. }
\begin{tabular}{c c c | c c c c c}
\hline
Data Aug & CLA Fusion & Loss Balance &PQ                   & $\text{PQ}^{\text{Th}}$ & $\text{PQ}^{\text{St}}$ &  SQ &  RQ  \\ \hline \hline
                &               &               & 35.4 & 40.5 & 27.7 & 77.2 & 43.5  \\ \hline
     $\surd $   &               &               & 37.4 & 43.2 & 28.6 & 77.6 & 45.7  \\ \hline
      $\surd $  &   $\surd $    &               & 38.4 & 43.0 & \textbf{31.4} & 77.7 & \textbf{47.6}  \\ \hline
    $\surd $    &   $\surd $    &   $\surd $    & \textbf{38.6} & \textbf{43.5} & 31.3 & \textbf{77.9} & 47.3  \\ \hline

\end{tabular}
\label{table:ablation}
\end{table}

\begin{table}[!tb]
\centering
     \caption{Ablation study on number of prototypes.}
    \begin{tabular}{l  | c c c c}
    \hline
    Prototypes &   PQ                   & $\text{PQ}^{\text{Th}}$ & $\text{PQ}^{\text{St}}$ & Inf time (ms) \\ \hline \hline
         16    & 38.4 & 43.4 & 30.8 & \textbf{50.7}\\ \hline
          32   & \textbf{38.6} & \textbf{43.5} & \textbf{31.3} & 51.1\\ \hline
        64     & 38.4 & 43.2 & 31.2  & 52.4 \\ \hline
    
    \end{tabular}
    \label{table:ablation_prototypes}
\end{table}

To verify the performance of training decisions and our cross-layer fusion attention module, we conduct the experiments with different settings in Table \ref{table:ablation} on COCO panoptic \emph{val} set. The empty cells in the table indicate that the corresponding component is not used.

\subsubsection{Data Augmentation.}
We compare the model without using data augmentation during training. The first and second rows in Table \ref{table:ablation} show that the model trained with data augmentation improves by 2\% in PQ. It proves that data augmentation plays important role during training.

\subsubsection{Cross-layer Attention Fusion.}
For the model without using cross-layer fusion module, it is replaced by another fusion module similar to Panoptic-FPN \cite{DeGeus2018a}. We adopt convolution blocks for different layers in backbone and combine them together. The layers $F_5$, $F_4$ and $F_3$ are attached with 3, 2 and 1 convolution blocks respectively and $2\times$ bilinear upsampling layers between each block. Output features are obtained by combining them with element-wise addition. As shown in second and third rows in Table \ref{table:ablation}, the model employing the cross-layer attention fusion module yields 38.4\% in PQ and 2.8\% improvement in PQ$^{\text{St}}$.

\subsubsection{Loss Balance.}
In order to balance the loss values in similar order of magnitude, we assign different weights for each loss during training. With loss balance, the weights are same as experimental setting. Without loss balance, all weights are set to 1.  As shown in the third and fourth rows in Table \ref{table:ablation}, the model with loss balance performs better especially on PQ$^{\text{Th}}$.

\subsubsection{Prototypes.}
We further conduct experiments on the number of prototypes. As shown in Table \ref{table:ablation_prototypes}, the number of prototypes barely influence the overall performance. However, it will affect the inference and post-processing time. We choose 32 prototypes for the setting of EPSNet. 

The ablation results show that our cross-layer attention fusion and training strategies bring significant improvement on panoptic segmentation.



\subsection{Analysis of Semantic Head}

\begin{table}[!tb]
    \begin{minipage}[t]{.45\textwidth}
    \centering
    \caption{Performance comparison on different design of semantic head. \emph{standard} denotes EPSNet with other design choice on semantic head, which directly generates semantic segmentation with convolutional layers. Note that, the EPSNet here does not use CLA fusion. M-adds denotes multiply-adds. } 
    \begin{tabular}{l  | c  c c c}
    \hline
    Method &   PQ                   & $\text{PQ}^{\text{Th}}$ & $\text{PQ}^{\text{St}}$  & M-Adds (M)\\ \hline \hline
         standard    & 37.2 & 42.9 & 28.5  &  33.8\\ \hline
        coefficients   & \textbf{37.4} & \textbf{43.2} & \textbf{28.6}  & \textbf{9.4}\\ \hline 
    
    \end{tabular}
    \label{table:semantic_head}
    \end{minipage}
    \hspace{.05\textwidth}
     \begin{minipage}[t]{.45\textwidth}
     \centering
     \caption{Performance comparison of using different options on semantic coefficients. } 
    \begin{tabular}{l  | c c c}
    \hline
    Method &   PQ                & $\text{PQ}^{\text{St}}$  \\ \hline \hline
         top-left     & 33.2 & 17.7  \\ \hline
        top-right    & 33.4 &  18.1   \\ \hline 
        bottom-left    & 33.5  &  18.3   \\ \hline 
        bottom-right    & 33.6 &  18.5   \\ \hline 
        center   & 37.5 &  28.5   \\ \hline  \hline
        max pooling    & 37.6 &  28.8   \\ \hline 
        average pooling    & \textbf{38.6} &  \textbf{31.3}   \\ \hline 
    
    \end{tabular}
    \label{table:sem_head_options}
    \end{minipage}
\end{table}

We further compare our semantic segmentation head to other design choice. In most proposed panoptic segmentation networks, they adopt feature maps in backbone and perform FCN \cite{Long2015} to obtain pixel-wise predictions. The semantic head of compared model is constructed by apply $1\times 1$ convolutional layer on the the fused feature maps with the Panoptic-FPN fusion, whose size is same as $F_3$. Note that we only replace the semantic head in EPSNet without CLA fusion. We count the multiply-adds to evaluate the computation cost of different structures for semantic heads. 

The experimental results in Table  \ref{table:semantic_head} show that the computation cost of the proposed semantic head is about 0.3 times less than the standard semantic head, although our semantic head is deeper. Unlike the standard semantic segmentation, because of the small input feature maps for computation, the proposed semantic head using mask coefficients does not slow down inference speed and outperforms the standard semantic head.

In semantic head, the coefficients $k\times N_{\text{stuff}}$ in each position is able to be used to generate semantic segmentation before average pooling. To verify the impact of coefficients from different position, we use the coefficients before average pooling from corner positions and center position to perform the semantic segmentations. In Table \ref{table:sem_head_options}, the comparison shows that the result using coefficients from center position is superior than other positions. Moreover, we compare the different options on pooling operation. The coefficients produced by average pooling yield better performance than using max pooling, as shown in Table \ref{table:sem_head_options}.

To sum up, the proposed semantic head predicts mask coefficients of each class with faster inference speed and efficiently exploits shared prototypes without dragging down panoptic segmentation performance. 

\subsection{Analysis of Cross-layer Attention Fusion Module }

\begin{figure}[!t]
\minipage{0.23\textwidth}
  \includegraphics[width=\linewidth]{images/cla_compare/000000022371.jpg}
\endminipage\hfill
\minipage{0.23\textwidth}
  \includegraphics[width=\linewidth]{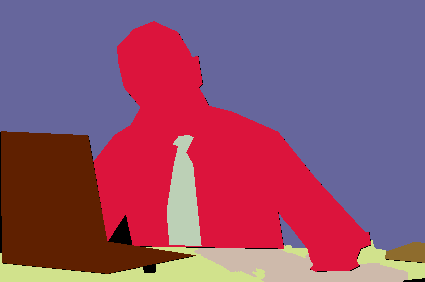}
\endminipage\hfill
\minipage{0.23\textwidth}%
  \includegraphics[width=\linewidth]{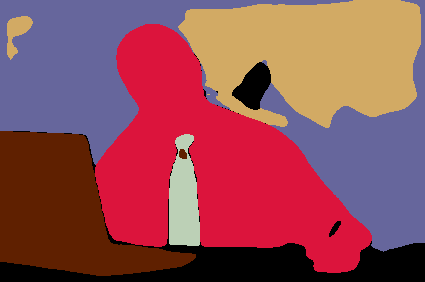}
\endminipage \hfill
\minipage{0.23\textwidth}%
  \includegraphics[width=\linewidth]{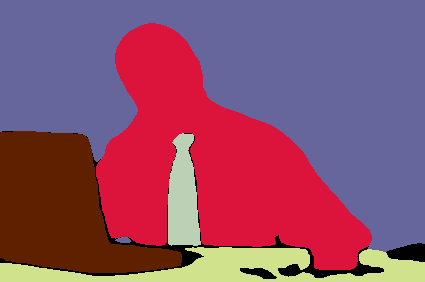}
\endminipage


\minipage{0.23\textwidth}
    \centering
  \includegraphics[width=\linewidth]{images/cla_compare/000000501005.jpg} 
  Input Image
\endminipage\hfill
\minipage{0.23\textwidth}
    \centering
  \includegraphics[width=\linewidth]{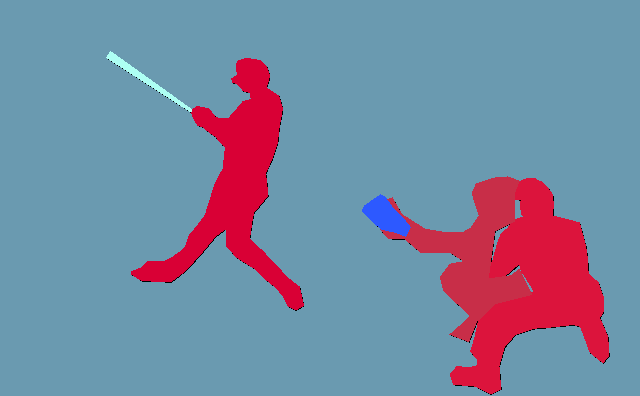}
  Ground Truth
\endminipage\hfill
\minipage{0.23\textwidth}%
    \centering
  \includegraphics[width=\linewidth]{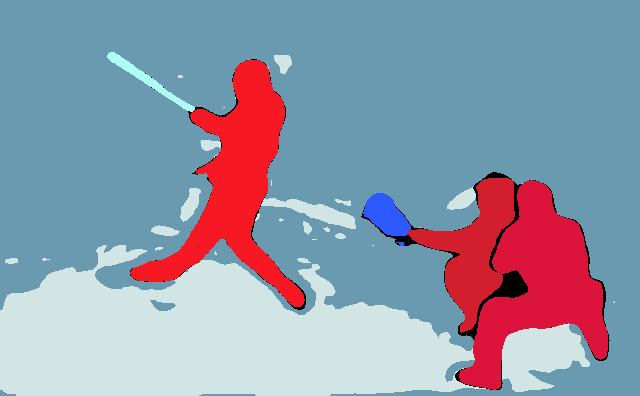}
  Normal Fusion
\endminipage \hfill
\minipage{0.23\textwidth}%
    \centering
  \includegraphics[width=\linewidth]{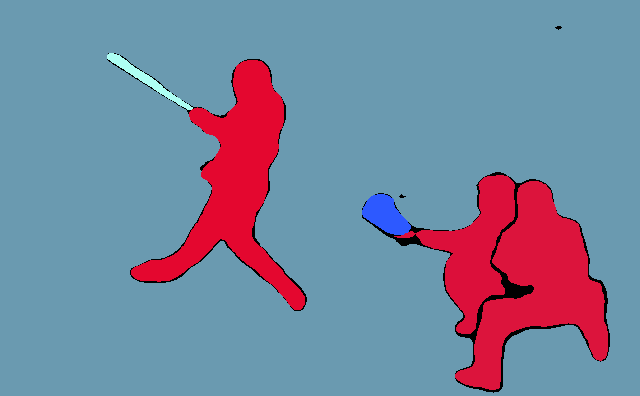}
  CLA Fusion
\endminipage
\caption{Visualization results of cross-layer attention fusion on COCO panoptic \emph{val} set. Normal fusion stands for the Panoptic-FPN  \cite{DeGeus2018a} fusion with $F_3$,$F_4$ and $F_5$. }
\label{fig:cla_fusion_compare}
\end{figure}

\begin{table}[!tb]
\centering
    \caption{Performance comparison on different strategies for fusion module. The inference time is measured without considering post-processing.}
        \begin{tabular}{l  | c c c c}
    \hline
    Method &   PQ                   & $\text{PQ}^{\text{Th}}$ & $\text{PQ}^{\text{St}}$ & Inf time \\ \hline \hline
         Panoptic-FPN fusion \cite{DeGeus2018a} ($F_3,F_4,F_5$)    & 37.4 & 43.2 & 28.6 & 23ms\\ \hline
         CLA fusion ($F_3$)    & 38.3 & 43.5 & 30.5  & 27ms\\ \hline
        CLA fusion ($F_3,F_4,F_5$)     & \textbf{38.6} & \textbf{43.5} & \textbf{31.3}  & 29ms \\ \hline
    
    \end{tabular}
    \label{table:cla_fusion}
\end{table}

In this subsection, we investigate different strategies of using cross-layer attention (CLA) fusion. We compare our EPSNet to the model using other fusion method like Panoptic-FPN \cite{DeGeus2018a} with $F_3$, $F_4$ and $F_5$ and another model only employing $F_3$ as source feature map with CLA fusion module. 
As shown in Table \ref{table:cla_fusion}, the model with CLA fusion outperforms the Panoptic-FPN fusion espescially on $\text{PQ}^{\text{St}}$. Also, more layers are adopted in CLA fusion can yield slight improvement and inference time.

The comparison of using cross-layer attention fusion can be visualized as Fig. \ref{fig:cla_fusion_compare}. The details of background are much better and clearer. CLA fusion helps model generate higher quality segmentation especially for the stuff classes. For instance, the segments of the table in the first row and the ground in the second rows are much more complete. 

\begin{figure}[t]
\small
\minipage{0.16\textwidth}
  \includegraphics[width=\linewidth, height=\linewidth]{images/attention/1-1.jpg}
\endminipage\hfill
\minipage{0.16\textwidth}
  \includegraphics[width=\linewidth]{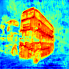}
\endminipage\hfill
\minipage{0.16\textwidth}%
  \includegraphics[width=\linewidth]{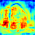}
\endminipage \hfill
\minipage{0.16\textwidth}
  \includegraphics[width=\linewidth, height=\linewidth]{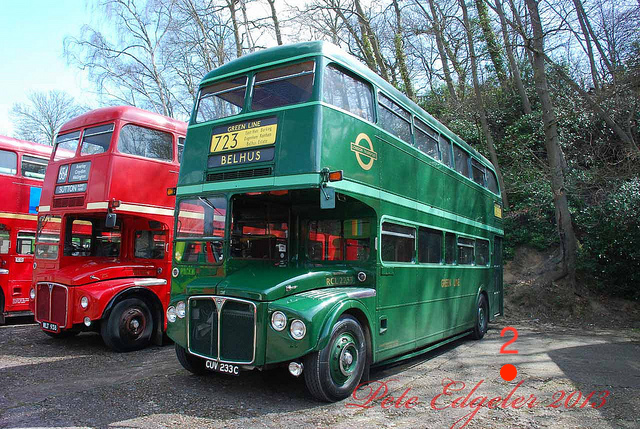}
\endminipage\hfill
\minipage{0.16\textwidth}
  \includegraphics[width=\linewidth]{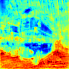}
\endminipage\hfill
\minipage{0.16\textwidth}%
  \includegraphics[width=\linewidth]{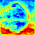}
\endminipage 


\minipage{0.16\textwidth}
\centering
  \includegraphics[width=\linewidth, height=\linewidth]{images/attention/3-1.jpg}
 Image \:    (point \#1) 
\endminipage\hfill
\minipage{0.16\textwidth}
\centering
  \includegraphics[width=\linewidth]{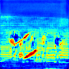}
  Sub-attention map ($F_3$)
\endminipage\hfill
\minipage{0.16\textwidth}%
\centering
  \includegraphics[width=\linewidth]{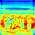}
  Sub-attention map ($F_4$)
\endminipage \hfill
\minipage{0.16\textwidth}
\centering
  \includegraphics[width=\linewidth, height=\linewidth]{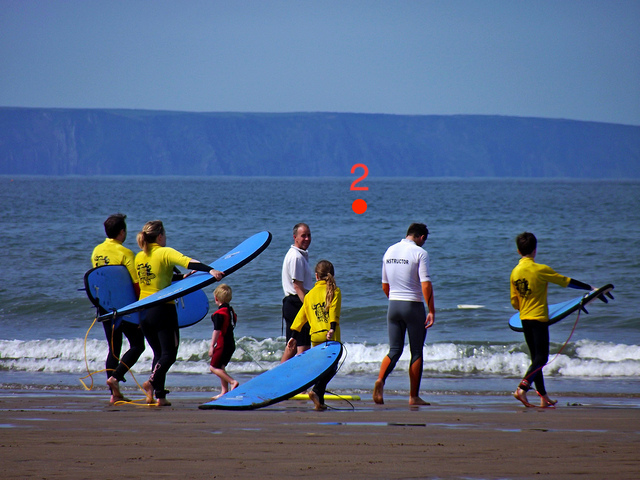}
 Image \:    (point \#2) 

\endminipage\hfill
\minipage{0.16\textwidth}
\centering
  \includegraphics[width=\linewidth]{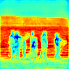}
  Sub-attention map ($F_3$)
\endminipage\hfill
\minipage{0.16\textwidth}%
\centering
  \includegraphics[width=\linewidth]{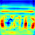}
  \small{Sub-attention map ($F_4$)}
\endminipage 
\caption{Visualization results of cross attention block on COCO panoptic \emph{val} set. In each row, we show the input images with different marked points in $1^{st}$ and $4^{th}$ columns and two sub-attention maps on source features ($F_3$ and $F_4$) corresponding to the marked point in  $2^{nd}$,  $3^{th}$,  $5^{th}$ and  $6^{th}$ columns.  }
\label{fig:cla_fusion_attention}
\end{figure}

To further understand what has been learned in CLA fusion module, we select two query points in input image in the first and fourth columns and visualize their corresponding sub-attention maps on other source features ($F_3$ and $F_4$) in remaining columns. In Fig. \ref{fig:cla_fusion_attention}, we observe that CLA fusion module can capture long-range dependencies according to the similarity. For example, in first row, the red point \#1 on bus pays more attention on  positions labeled as bus (second and third columns). For the point \# 2 on ground, it highlights most areas labeled as ground(fifth and sixth columns).

\subsection{Comparison with Other Methods on COCO}
 
\begin{table}[tb]
\small
\centering
\caption{Panoptic segmentation results on COCO \emph{val} set. \emph{LW-MNV2} denotes Light Wider MobileNet-V2. \emph{W-MNV2} means Wider MobileNet-V2. }
\begin{tabular}{l | c | c| c c  c | c}
\hline
Method    & Backbone & Input Size &PQ            & $\text{PQ}^{\text{Th}}$ & $\text{PQ}^{\text{St}}$ & Inf time (ms) \\ \hline \hline
 \multicolumn{7}{c}{\text{Two Stage} } \\ \hline

JSIS \cite{DeGeus2018}  &  ResNet-50      & $400\times 400$   & 26.9           &  29.3 &  23.3  &  -         \\ 

AUNet \cite{Li2018c} &  ResNet-50-FPN      &  -  & 39.6            &  49.1 &  25.2  &   -     \\ 

Panoptic-FPN \cite{DeGeus2018a}  &  ResNet-101-FPN     &  -  &  40.3         &  47.5 &  29.5  &  -       \\ 
AdaptIS \cite{Sofiiuk2019} &   ResNeXt-101      &  -  & 42.3          &  49.2 &  31.8  &    -   \\
UPSNet \cite{Xiong2019} &  ResNet-50-FPN      & $800\times 1333$   & 42.5            &  48.6 &  33.4  & 167       \\ \hline

 \multicolumn{7}{c}{\text{Single Stage} } \\ \hline
DeeperLab \cite{Yang2019} &  LW-MNV2         & $641\times641$   & 24.1             &   -    &    -   & 73      \\
DeeperLab \cite{Yang2019} &  W-MNV2         & $641\times641$   & 27.9             &   -   & -      & 83       \\
DeeperLab \cite{Yang2019} &  Xception-71         & $641\times641$   & 33.8             &   -     &    -   & 119       \\

SSAP \cite{Porzi2019} &  ResNet-101            & $512\times512$   & 36.5             &   -     &    - &     -   \\ \hline
Ours      &  ResNet-101-FPN  & $550\times550$   & $\textbf{38.6}$  &  43.5  &  31.3 &  \textbf{51} \\ \hline
\end{tabular}
\label{table:COCO_val_results}
\end{table}

\begin{table}[tb]
\small
\centering
\caption{Panoptic segmentation results on COCO \emph{test-dev} set. \emph{LW-MNV2} denotes Light Wider MobileNet-V2. \emph{W-MNV2} means Wider MobileNet-V2. \emph{Flip} and \emph{MS} stands for horizontal flipping and multi-scale inputs during testing.}
\begin{tabular}{l | c | c c | c c c}
\hline
Method    & Backbone & Flip & MS & PQ   & $\text{PQ}^{\text{Th}}$ & $\text{PQ}^{\text{St}}$ \\ \hline \hline
 \multicolumn{7}{c}{\text{Two Stage} } \\ \hline

JSIS \cite{DeGeus2018}  &  ResNet-50      &   &   & 27.2           &  29.6 &  23.4         \\ 
Panoptic-FPN \cite{DeGeus2018a}  &  ResNet-101-FPN     &   &   &  40.9         &  48.3 &  29.7       \\ 

AdaptIS \cite{Sofiiuk2019} &   ResNeXt-101      &  \checkmark &   & 42.8          &  50.1 &  31.8   \\
AUNet \cite{Li2018c} &  ResNeXt-101-FPN      &    &    \checkmark  & 46.5           &  55.8 &  32.5     \\ 
UPSNet \cite{Xiong2019} &  ResNet-101-FPN      &  \checkmark &  \checkmark   & 46.6            &  53.2 &  36.7        \\ \hline

 \multicolumn{7}{c}{\text{Single Stage} } \\ \hline
DeeperLab \cite{Yang2019} &  Xception-71         &   &    & 34.3             &   37.5     &    29.6          \\

SSAP \cite{Porzi2019} &  ResNet-101            &  \checkmark &   \checkmark & 36.9            &  40.1 &   \textbf{32.0}    \\ \hline
Ours      &  ResNet-101-FPN  & &    & $\textbf{38.9}$  &  \textbf{44.1}  &  31.0  \\ \hline
\end{tabular}
\label{table:COCO_test_results}
\end{table}
We compare our method on COCO  $val$ set with panoptic quality and inference speed measured  from input image to panoptic segmentation  output including post-processing time. Specifically, our model is only trained on COCO training dataset with ResNet-101-FPN and tested using single-scale $550\times550$ image. As shown in Table \ref{table:COCO_val_results}, EPSNet outperforms every one-stage method and improves the performance over DeeperLab \cite{Yang2019} with Xception-71 \cite{Chollet2017} backbone by 4.8\% PQ. Also, our inference speed is much faster than all existing panoptic segmentation methods. EPSNet only takes 51ms for inference, which is $1.4\times$ faster than DeeperLab with Light Wider MobileNet-V2 \cite{Sandler2018} backbone and $3.1\times$ faster than UPSNet \cite{Xiong2019}. Compared to the two-stage methods,  we bring better performance especially on PQ$^{\text{St}}$, which outperforms Panoptic-FPN \cite{DeGeus2018a} by 1.8\%, indicating that our approach provides better results on semantic segmentation. 
Despite the one-stage detector of EPSNet, with the fusion module and efficient architecture, we not only achieve competitive result for panoptic segmentation but also boost the inference speed. 

In COCO \emph{test} set, the inference setting is the same as COCO \emph{val} set experiment. As shown in Table \ref{table:COCO_test_results}, we outperform SSAP \cite{Porzi2019}, which adopts horizontal flipping and multi-scale input images for testing , by 2\% PQ. Without any additional tricks for inference, we still achieve competitive result compared to other methods.

\section{Conclusions}
In this paper, we present a one-stage Efficient Panoptic Segmentation Network. The masks are efficiently constructed by linear combination of prototypes generated by protohead and mask coefficients produced by instance and semantic branches. The proposed cross-layer attention fusion module aggregates multi-scale features in different layers with attention mechanism to enhance the quality of shared prototypes. The experiments show that our method achieves competitive performance on COCO panoptic dataset and outperforms other one-stage approaches. Also, EPSNet is significantly faster than the existing panoptic segmentation networks. 
In the future, We would like to explore a more effective way to replace the heuristic merging algorithm.



\bibliographystyle{splncs}

\bibliography{thesis}

\end{document}